\documentclass{article}

\usepackage{microtype}
\usepackage{graphicx}
\usepackage{subfigure}
\usepackage{booktabs} 
\usepackage{multicol}

\usepackage{hyperref}

\usepackage{mathtools}

\usepackage[accepted]{icml2021}

\icmltitlerunning{Evaluating subgroup disparity using epistemic uncertainty in mammography}

\begin{document}

\twocolumn[
\icmltitle{Evaluating subgroup disparity using epistemic uncertainty in mammography}



\icmlsetsymbol{equal}{*}

\begin{icmlauthorlist}
\icmlauthor{Charles Lu}{mgh}
\icmlauthor{Andreanne Lemay}{mgh}
\icmlauthor{Katharina Hoebel}{mgh,mit}
\icmlauthor{Jayashree Kalpathy-Cramer}{mgh}
\end{icmlauthorlist}

\icmlaffiliation{mgh}{Massachusetts General Hospital}
\icmlaffiliation{mit}{Massachusetts Institute of Technology}

\icmlcorrespondingauthor{Charles Lu}{clu@mgh.harvard.edu}

\icmlkeywords{deep learning, uncertainty, fairness, mammography}

\vskip 0.3in
]



\printAffiliationsAndNotice{}  

\begin{abstract}
    As machine learning (ML) continue to be integrated into healthcare systems that affect clinical decision making, new strategies will need to be incorporated in order to effectively detect and evaluate subgroup disparities to ensure accountability and generalizability in clinical workflows.
    In this paper, we explore how epistemic uncertainty can be used to evaluate disparity in patient demographics (race) and data acquisition (scanner) subgroups for breast density assessment on a dataset of 108,190 mammograms collected from 33 clinical sites.
    Our results show that even if aggregate performance is comparable, the choice of uncertainty quantification metric can significantly the subgroup level.
    We hope this analysis can promote further work on how uncertainty can be leveraged to increase transparency of machine learning applications for clinical deployment.
\end{abstract}

\section{Introduction}
\label{intro}
    Machine learning holds tremendous promise for applications in healthcare yet real world deployment remains challenging due to difficulties in developing robust and interpretable models suitable for integration into existing clinical systems.
    Issues of algorithmic bias in ML have been extensively studied, several formal definitions of fairness proposed \cite{DBLP:journals/corr/LouizosSLWZ15, NIPS2016_9d268236, NIPS2017_a486cd07}.
    Previous works demonstrates how proxy variables and black box models can exacerbate biases in several different healthcare applications \cite{Buolamwini2018GenderClassification, Chen2019CanCare, Obermeyer447, Larrazabal2020GenderDiagnosis, 2020arXiv200300827S}.
    The potential for learning systems to significantly increase disparity in protected subgroups would be a major limitation to widespread deployment of ML into critical clinical infrastructure \cite{Du2020FairnessPerspective}.

    For ML to be adopted in healthcare, new strategies and perspectives will need to be developed and incorporated to better monitor and audit algorithmic-assisted workflows for robustness and transparency. 
    Inspired by works, such as \citet{Bhatt2020UncertaintyUncertainty}, which advocates for the use of uncertainty as one approach to achieve more transparency in ML, and \citet{Dusenberry2020AnalyzingRecords}, which shows ML model exhibiting higher uncertainty for different demographics in electronic health records, we investigate how uncertainty could be used to detect and evaluate disparities between subgroups.
    Subgroup analysis is important in many clinical settings, especially in non-stationary distributions where characteristics of data acquisition change over time (e.g. clinical equipment is upgraded or protocols are changed)\cite{DBLP:journals/corr/abs-1806-00388}.
    The integration of predictive uncertainty can have implications for both regulation and generalizability of machine learning applications in healthcare by allowing clinicians more easily interpret predictions, investigate possible failure modes, and adjust safety thresholds according to different clinical contexts.
    Incorporating uncertainty thresholds into clinical ML workflows can be set to abnormal and ambiguous cases for manual review and to monitor how models behave under distribution shifts or on subgroups with different base rates.
    Uncertainty modeling can be broadly categorized into \textit{epistemic} (model uncertainty) and \textit{aleatoric} (data uncertainty)\cite{Kiureghian2009AleatoryMatter}.
    Epistemic uncertainty is important in many data-limited and safety-critical applications, such as medical image analysis, and can indicate poor generalizability \cite{Kendall2017WhatVision, DeVries2018LeveragingQuality}.

    Popular Bayesian approximation methods used to assess epistemic uncertainty include Deep Ensembles \cite{Lakshminarayanan2017SimpleEnsembles} and Monte Carlo (MC) Dropout \cite{Gal2016DropoutGhahramani}. 
    \citet{Leibig2017LeveragingDetection} showed excluding samples with high uncertainty, which could be referred later for additional followup and review, overall performance improved in diabetic retinopathy screening.
    Similarly, in the classification of dermoscopic images, epistemic uncertainty was higher for misclassified and out-of-distribution data \cite{Combalia2020UncertaintyClassification}.

    In this paper, we evaluate several common forms of uncertainty quantification (UQ) on a large, heterogeneous mammography dataset for breast density assessment and propose a novel disparity metric based on epistemic uncertainty to show how disparities manifest in racial demographics and scanner acquisition subgroups.
    This metric may be useful in characterizing failure modes or detecting distribution shift.

\section{Methods}
\label{methods}
    \subsection{Measures of uncertainty quantification}
        We use Monte Carlo dropout \cite{Gal2016DropoutGhahramani} as a Bayesian approximation to epistemic uncertainty to sample $T$ softmax probabilities, $\{p_t\}^T_{t=1} = p_1, p_2, \ldots, p_t$, which are averaged along the class dimension for the final prediction, $\hat{y} = \underset{C}{\arg \max}\frac{1}{T}\sum_{t=1}^T p_{t \mid c}$.
        We compare the following uncertainty metrics: maximum softmax probability, predictive variance, predicted entropy, and Bhattacharyya coefficient.

    As detailed in \citet{hendrycks17baseline} to detect out-of-distribution examples, we use maximum softmax probability (Naive) as our baseline uncertainty.
    \begin{equation}
        \mathrm{Naive}(p) = 1 - \max(p)
    \end{equation}

    \textbf{Predictive variance} is defined as the average variance computed over $T$ MC samples
    \begin{equation}
        \mathrm{Var}\left(\{p\}_{t=1}^T\right) = \frac{1}{C} \sum_{c=1}^C \mathrm{Var}(p_c)
    \end{equation}
    where $Var(p_c) = \frac{1}{T} \sum_{t=1}^T \left(p_{t \mid c} - \frac{1}{T} \sum_{t=1}^T p_{t \mid c}] \right)^2$.

    \textbf{Predictive entropy} the expected information over $T$ MC samples
    \begin{equation}
        \mathrm{H}\left(\{p\}_{t=1}^T\right) = -\frac{1}{C} \sum_{c=1}^C \left(\frac{1}{T} \sum_{t=1}^T p_{t \mid c} \log \frac{1}{T} \sum_{t=1}^T p_{t \mid c}\right)
    \end{equation}

    \textbf{Bhattacharyya coefficient} (BC) is a measure of similarity between two samples. 
    In this case, we compute BC using $N$-bin histogram of the top two classes, $h^1 = \mathrm{histogram}(\{p_{c=c_1}\}_{t=1}^T$ and $h^2 = \mathrm{histogram}(\{p_{c=c_2}\}_{t=1}^T$, as determined by their predictive mean.
    \begin{equation}
        \textrm{BC}\left(\{p\}_{t=1}^T\right) = \frac{1}{N} \sum_{n=1}^N \sqrt{h_n^1 h_n^2}
    \end{equation}

    Subsequently, given some measure of uncertainty quantification $\mathrm{UQ}$, a performance metric $\mathrm{PM}(\hat{Y}, Y)$, a set of rejection thresholds, $\{r\}_{i=1}^n = r_1, r_2, \ldots, r_n$, and a mutually exclusive subgroup attribute, $\mathcal{A}$, such as patient race or scanner model, we can define the following subgroup disparity metric
    \begin{equation}
        \delta = \frac{1}{n} \sum_{i=1}^n \sum_{A, A^\prime \in {\mathcal{A} \choose 2}} \bigg\vert f(r_i, A) - f(r_i, {A^\prime}) \bigg\vert 
    \end{equation}
    where $\mathcal{A} \choose 2$ is the set of unique pairwise subgroups and $f = \mathrm{PM}\left(\{\hat{y} \in \hat{Y} \mid A,\; \mathrm{UQ}(\hat{y}) < r_i\}, \{y \in Y \mid A,\; \mathrm{UQ}(\hat{y}) < r_i\}\right)$ is the performance on subgroup $A$ at threshold $r_i$.

    \subsection{Dataset}
    We perform all experiments on the Digital Mammographic Imaging Screening Trial (DMIST) dataset, which comprises 108,230 mammograms from 21,729 patients from 33 clinical sites and interpreted by a total of 92 radiologists \cite{Pisano2005DiagnosticScreening}.
    Accurate assessment of breast density, as measured by relative amounts of fibroglandular tissue, is important for the interpretation of mammographies and is commonly rated according to the BI-RADS criteria into one of four categories: entirely fatty, scattered, heterogeneously dense, and extremely dense \cite{bi-rads}.


    \subsection{Experiments}
    To assess the performance of breast density task, we use linearly weighted Cohen's Kappa score, $\kappa \in [-1, 1]$, which is commonly used to measure agreement between two raters, where $P_o$ is the proportion of observed agreement and $P_e$ is the agreement to due random chance \cite{kappa}.
    \begin{equation}
        \kappa = \frac{P_o - P_e}{1 - P_e}
    \end{equation}
    We consider two subgroups to analyze on a patient level: 5 races (white, black, hispanic, asian, and other) and and 4 scanner types (Senograph, Senscan, ADS, and other).

    We split the DMIST dataset by patients into a held-out test set of 38,936 (35.9\%) mammograms and a development set of 69,254 mammograms, which was further split randomly into training and validation sets (90\% / 10\% respectively).
    All experiments used the ResNet50 \cite{He2016DeepRecognition} architecture with dropout layers ($p = 0.1$), implemented using the MONAI framework \cite{Ma2021Project-MONAI/MONAI:0.5.3}, finetuned from ImageNet \cite{Russakovsky2015ImageNetChallenge}, and trained with cross-entropy loss using the ADAM optimizer for up to $100$ epochs with early stopping using patience of $5$ epochs.
    Random data augmentation in rotations, flipping, resizes, affine transforms, and cropping was applied during training.

    Each experiment was performed five times with different random seeds.
    To estimate uncertainty, $100$ Monte Carlo predictions were sampled for each test image and final predictions were obtained by averaging over all samples.
    Finally, predictions from distinct mammograms of the same patient were averaged to obtain patient level predictions.

\section{Results}
\label{results}
    The overall Kappa score on our test set was $0.646 \pm 0.016$.
    Unsurprisingly, performance is highest for the most represented subgroups in the data distribution: $0.647 \pm 0.015$ for white patients and $0.658 \pm 0.021$ Senograph scanners (Table \ref{tab:main}).
    Comparing within subgroups, we do not find any significant relationship between uncertainty and race and a moderate relationship between number of examples and uncertainty for acquisition type (Table \ref{tab:main}).
    For breast density, there is a strong negative correlation between class proportion and uncertainty, with scattered and heterogeneous having lower uncertainty than fatty and dense classes (Table \ref{tab:main}).

    \begin{figure}[ht]
    \vskip 0.2in
    \begin{center}
    \centerline{\includegraphics[width=\columnwidth]{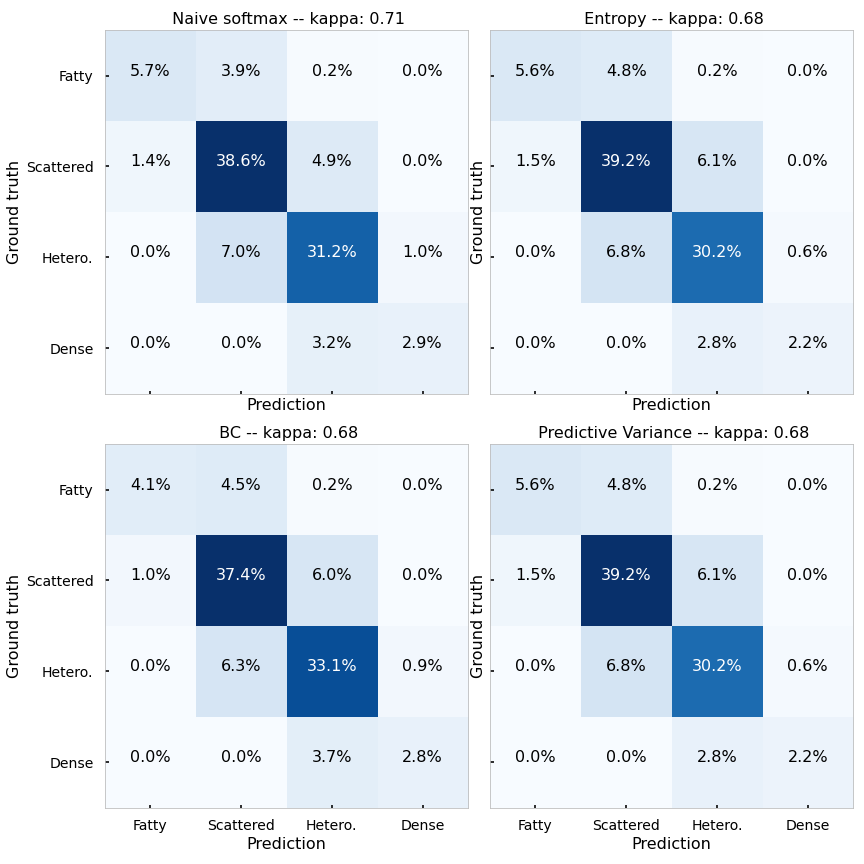}}
    \caption{Confusion matrix of four types of uncertainty measures (maximum softmax probability, predictive entropy, Bhattacharyya coefficient, and predictive variance) at 20\% uncertainty threshold}
    \label{fig:confusion}
    \end{center}
    \vskip -0.2in
    \end{figure}

\begin{figure}[ht]
    \vskip 0.2in
    \begin{center}
    \centerline{\includegraphics[width=\columnwidth]{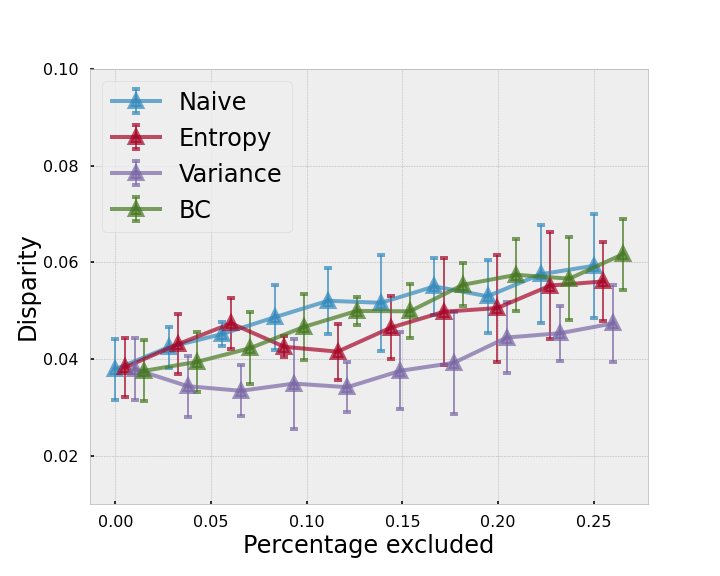}}
    \caption{Average Kappa disparity between race subgroup as uncertainty threshold increases (lower disparity is better).}
    \label{fig:race}
    \end{center}
    \vskip -0.2in
    \end{figure}

\begin{figure}[ht]
    \vskip 0.2in
    \begin{center}
    \centerline{\includegraphics[width=\columnwidth]{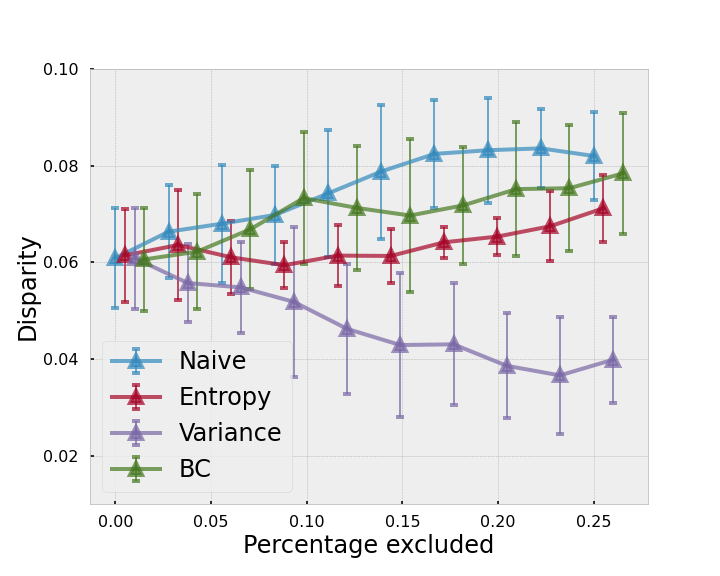}}
    \caption{Average Kappa disparity between scanner subgroup as uncertainty threshold increases (lower disparity is better).}
    \label{fig:scan}
    \end{center}
    \vskip -0.2in
    \end{figure}

\begin{figure}[ht]
    \vskip 0.2in
    \begin{center}
    \centerline{\includegraphics[width=\columnwidth]{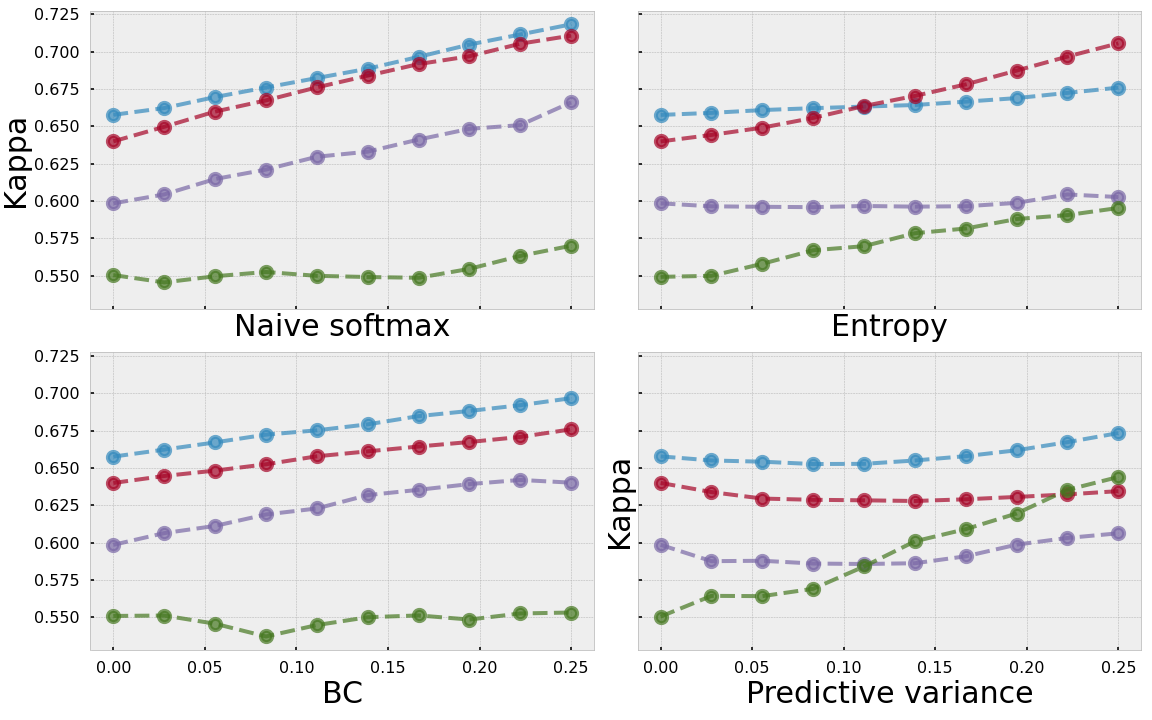}}
    \caption{Comparison of uncertainty performance by acquisition scanner type (blue is Senograph; red is Senoscan; purple is ADS; green is Other)}
    \label{fig:scanline}
    \end{center}
    \vskip -0.2in
    \end{figure}

    As shown in Figure \ref{fig:confusion}, aggregate Kappa score is comparable between uncertainty metrics with similar distribution of breast density ratings at 20\% uncertainty threshold.
    However, when analyzing at subgroup level, we observe that predictive variance has significantly lower disparity for both race and scanner categories (Figures \ref{fig:race} - \ref{fig:scan} and Table \ref{tab:sub}).
    In Figure \ref{fig:race}, we see increasing trend in subgroup disparity, even as overall performance improves as more uncertain cases are rejected, with predictive variance remaining the most stable across the range of considered rejection thresholds.
    For scanner subgroups in Figure \ref{fig:scan}, we observe that predictive entropy and predictive variance to be relatively stable for low uncertainty thresholds and predictive variance significantly decreasing in disparity at higher thresholds 25\%. The race and acquisition disparities at different rejection thresholds are also listed in Table \ref{tab:sub}.
    Further examining the difference in Kappa scores between scanner acquisition types in Figure \ref{fig:scanline}, we observe that predictive variance performed especially well on the least represented scanner type.

    %

\begin{table*}[t]
\caption{Performance (mean $\pm$ std), subgroup disparity, and distribution by subgroup (race, scanner, label). Note that kappa score is not well defined on a single class label for breast density subgroups.}
\label{tab:main}
\vskip 0.15in
\begin{center}
\begin{small}
\begin{sc}
\begin{tabular}{lcccccc}
\toprule
 & $\kappa$ & Naive & Var. (\small{$\times 10^{-3}$}) & Entr. & BC  & Distribution\\
\midrule
\bf{Race} \\
\midrule
White    & \bf{0.647 $\pm$ 0.015} & 0.287 $\pm$ 0.004 & \bf{3.5 $\pm$ 0.2} & \bf{0.172 $\pm$ 0.003} & 0.808 $\pm$ 0.005 & 80.7\% \\
Black & 0.641 $\pm$ 0.019 & 0.292 $\pm$ 0.004 & 3.9 $\pm$ 0.3 & 0.173 $\pm$ 0.003 & 0.808 $\pm$ 0.004 & 13.1\% \\
Hisp. & 0.573 $\pm$ 0.020 & \bf{0.285 $\pm$ 0.007} & 4.3$\pm$ 0.4 & 0.175 $\pm$ 0.006 &  \bf{0.795 $\pm$ 0.010} & 3.7\% \\
Asian   & 0.627 $\pm$ 0.042 & 0.296 $\pm$ 0.006 & \bf{3.5$\pm$ 0.2} & 0.180 $\pm$ 0.004 &  0.809 $\pm$ 0.009 & 1.8\% \\
Other    & 0.625 $\pm$ 0.031 & 0.286 $\pm$ 0.007 & 3.7$\pm$ 0.1 & 0.174 $\pm$ 0.003 &  \bf{0.795 $\pm$ 0.010} & 0.7\% \\
\midrule
\multicolumn{5}{l}{\bf{Acquisition type}} \\
\midrule
Senograph  & \bf{0.658 $\pm$ 0.021} & 0.280 $\pm$ 0.004 & 3.4 $\pm$ 0.3 & 0.166 $\pm$ 0.003 &  0.801 $\pm$ 0.004 & 55.8\% \\
Senoscan & 0.641 $\pm$ 0.011 & 0.304 $\pm$ 0.005 & 3.7 $\pm$ 0.2 & 0.183 $\pm$ 0.003 &  0.822 $\pm$ 0.007 & 36.4\% \\
Ads & 0.599 $\pm$ 0.024 & 0.272 $\pm$ 0.005 & \bf{3.2 $\pm$ 0.3} & \bf{0.164 $\pm$ 0.004} & 0.793 $\pm$ 0.003 & 5.4\% \\
Other   & 0.551 $\pm$ 0.025 & \bf{0.270 $\pm$ 0.012} & 5.9 $\pm$ 0.5 & 0.174 $\pm$ 0.010 &  \bf{0.764 $\pm$ 0.018} & 2.5\% \\
\midrule
\multicolumn{5}{l}{\bf{Breast density}} \\
\midrule
Fatty  & N/A & 0.301 $\pm$ 0.008 & 4.3 $\pm$ 0.6 & 0.172 $\pm$ 0.002 &  0.840 $\pm$ 0.011 & 10.8\% \\
Scattered & N/A & \bf{0.279 $\pm$ 0.008} & \bf{3.3 $\pm$ 0.3} & \bf{0.171 $\pm$ 0.004} &  0.805 $\pm$ 0.008 & 43.2\% \\
Heterogeneous & N/A & 0.291 $\pm$ 0.008 & 3.4 $\pm$ 0.2 & 0.173 $\pm$ 0.004 & \bf{0.801 $\pm$ 0.006} & 38.9\% \\
Dense & N/A & 0.307 $\pm$ 0.011 & 5.0 $\pm$ 0.4 & 0.178 $\pm$ 0.006 &  0.806 $\pm$ 0.012 & 7.1\% \\
\midrule
Overall & 0.646 $\pm$ 0.016 & 0.288 $\pm$ 0.004 &  3.6 $\pm$ 0.2 & 0.172 $\pm$ 0.003  &  0.807 $\pm$ 0.004 & \\
\bottomrule
\end{tabular}
\end{sc}
\end{small}
\end{center}
\vskip -0.1in
\end{table*}

\begin{table}[t]
\caption{Change in disparity (mean $\pm$ std) by uncertainty metric for racial and acquisition subgroups at different rejection thresholds.}
\label{tab:sub}
\vskip 0.15in
\begin{center}
\begin{small}
\begin{sc}
\begin{tabular}{lccc}
\toprule
& \multicolumn{3}{c}{Proportion of excluded cases} \\
\midrule
& \bf{1\%} & \bf{10\%} & \bf{25\%} \\
\midrule
\multicolumn{4}{l}{\bf{Race disparity}} \\
\midrule
Naive & 0.041 $\pm$ 0.005 & 0.051 $\pm$ 0.007 & 0.059 $\pm$ 0.011 \\
Var.    & 0.037 $\pm$ 0.008  & 0.034 $\pm$ 0.007 & 0.047 $\pm$ 0.008 \\
Entr. & 0.040 $\pm$ 0.007 & 0.040 $\pm$ 0.004 & 0.056 $\pm$ 0.008\\
BC & 0.038 $\pm$ 0.005 & 0.048 $\pm$ 0.003 & 0.062 $\pm$ 0.007 \\
\midrule
\multicolumn{4}{l}{\bf{Acquisition disparity}} \\
\midrule
Naive & 0.064 $\pm$ 0.010 & 0.073 $\pm$ 0.013 & 0.082 $\pm$ 0.009 \\
Var.    & 0.061 $\pm$ 0.012 & 0.050 $\pm$ 0.015 & 0.040 $\pm$ 0.009 \\
Entr. & 0.063 $\pm$ 0.011 & 0.062 $\pm$ 0.006 & 0.071 $\pm$ 0.007 \\
BC & 0.061 $\pm$ 0.012 & 0.070 $\pm$ 0.013 & 0.078 $\pm$ 0.012 \\
\midrule
Kappa $\kappa$ & 0.646 $\pm$ 0.001 & 0.657 $\pm$ 0.012 & 0.682 $\pm$ 0.019 \\
\bottomrule
\end{tabular}
\end{sc}
\end{small}
\end{center}
\vskip -0.1in
\end{table}


\section{Conclusion}
\label{conclusion}
    In this work, we proposed a simple evaluation strategy to analyze subgroup disparities to compare different measures of epistemic uncertainty.
    Our experiments on a large dataset of mammograms demonstrate how the choice of uncertainty metric can disproportionately affect subgroups, such as race and scanner, even when aggregate performance appear similar.
    We find that using predictive variance for uncertainty maintained the lowest level of disparity in both race and scanner subgroups at different rejection thresholds.
    Interestingly, disparity in scanner actually decreased substantially at higher thresholds with predictive variance.
    Future work might explore subgroup uncertainty in other clinical datasets as well as evaluating other definitions of uncertainty in healthcare contexts.
    Additionally, human reader studies will be crucial in investigating how clinical users interface with uncertainty estimates in different contexts to make a decision.

%
%




%

\bibliography{references}
\bibliographystyle{icml2021}


\end{document}